\documentclass[runningheads]{llncs}
\usepackage{graphicx}
\usepackage{amsmath,amssymb} 
\usepackage{color}

\newcommand{\pa}[1]{\left(#1\right)} 

\newcommand{\abs}[1]{\left\vert #1 \right\vert}     

\definecolor{col_a}{RGB}{178,24,43}
\definecolor{col_b}{RGB}{10,10,10}
\definecolor{col_c}{RGB}{31,120,180}
\definecolor{col_d}{RGB}{51,160,44}
\definecolor{col_e}{RGB}{30,30,30}
\definecolor{col_f}{RGB}{99,99,99}
\definecolor{col_g}{RGB}{118,42,131}
\definecolor{col_h}{RGB}{140,81,10}

\begin{document}

\newcommand{\point}{
    \raise0.7ex\hbox{.}
    }


\pagestyle{headings}

\mainmatter

\title{End-to-End Learning for \protect\\ Image Burst Deblurring} 

\titlerunning{End-to-End Learning for Image Burst Deblurring} 

\authorrunning{P. Wieschollek \and B. Sch\"olkopf \and H. P.A. Lensch \and M. Hirsch } 

\author{Patrick Wieschollek$^{1,2}$ \and Bernhard Sch\"olkopf$^{1}$  \and \protect\\ Hendrik P.A. Lensch$^{2}$ \and Michael Hirsch$^{1}$ } 
\institute{${}^1$ Max Planck Institute
for Intelligent Systems, ${}^2$ University of T\"ubingen} 

\maketitle

\begin{abstract}

We present a neural network model approach for multi-frame blind
deconvolution. The discriminative approach adopts and combines two
recent techniques for image deblurring into a single neural network
architecture.  Our proposed hybrid-architecture combines the explicit
prediction of a deconvolution filter and non-trivial averaging of
Fourier coefficients in the frequency domain.  In order to make full
use of the information contained in all images in one burst, the
proposed network embeds smaller networks, which explicitly allow the
model to transfer information between images in early layers. Our
system is trained end-to-end using standard backpropagation on a set
of artificially generated training examples, enabling competitive
performance in multi-frame blind deconvolution, both with respect to quality and
runtime.
\end{abstract}


\section{Introduction}
\label{sec:introduction}

Nowadays, consumer cameras are able to capture an entire series of
photographs in rapid succession.  Hand-held acquisition of a burst of
images is likely to cause blur due to unwanted camera shake during
image capture. This is particularly true along with longer exposure
times needed in low-light environments.

Motion blurring due to camera shake is commonly modeled as a spatially
invariant convolution of a latent sharp image $X$ with an unkown blur
kernel $k$
\begin{equation}
  Y = k * X + \varepsilon,
  \label{eq:model}
\end{equation}
where $*$ denotes the convolution operator, $Y$ the blurred
observation and $\varepsilon$ additive noise. Single image blind
deconvolution (BD), i.e. recovering $X$ from $Y$ without knowing $k$,
is a highly ill-posed problem for a variety of reasons. In contrast,
multi-frame blind deconvolution  or burst deblurring methods aim
at recovering a single sharp high-quality image from a sequence of
blurry and noisy observed images $Y_1, Y_2, \ldots, Y_N$. Accumulating
information from several observations can help to solve the
reconstruction problem associated with Eq.\eqref{eq:model} more
effectively.

Traditionally, generative models are used for blind image
deconvolution. While they offer much flexibility they are often
computationally demanding and time-consuming.

Discriminative approaches on the other hand keep the promise of fast
processing times, and are particularly suited for situations where an
exact modeling of the image formation process is not possible. A
popular choice in this context are neural networks.
They gained some momentum due to the great success of deep learning in many
supervised computer vision tasks, but also for a number of low-level
vision tasks state-of-the-art results have been reported
\cite{burger2012image,schuler2013machine}.  Our proposed method lines
up with the latter approaches and comes along with the following main
contributions:
\begin{enumerate}
  \item A robust state-of-the-art method for multi-image
    blind-deconvolution for both invariant and spatially-varying blur.
  \item A hybrid neural network architecture as a discriminative approach for image deblurring supporting end-to-end learning in the fashion of deep learning. 
  \item A neural network layer version of Fourier-Burst-Accumulation \cite{fba} with learnable weights.
  \item The proposed embedding of a small neural network allowing for information sharing accross the image burst early in the processing stage.
\end{enumerate}

\section{Related Work}
\label{sec:related_work}

Blind image deconvolution (BD) has seen considerable progress in the
last decade. A comprehensive review is provided in the recent overview
article by Wang and Tao \cite{wang2014recent}. 

\textbf{Single image blind deconvolution.} Approaches for single image
BD that report state-of-the-art results include the methods of Sun et
al. \cite{sun2013edge}, and Michaeli and Irani
\cite{michaeli2014blind} that use powerful patch-based priors for
image prediction. Following the success of deep learning methods in
computer vision, also a number of neural network based methods have
been proposed for image restoration tasks including \emph{non-blind}
deconvolution
\cite{schuler2013machine,convnetdeblur,rosenbaum2015return} which
seeks to restore a blurred image when the blur kernel is known, but
also for the more challenging task of \emph{blind} deconvolution
\cite{Schuler_PAMI15,sun2015learning,chakrabarti,hradivs2015convolutional,svoboda2016cnn,loktyushin2015retrospective}
where the blur kernel is not known a priori. Most relevant to our work
is the recent work of Chakrabarti \cite{chakrabarti} which proposes a
neural network that is trained to output the complex Fourier
coefficients of a deconvolution filter. When applied to an input patch
in the frequency domain, the network returns a prediction of the
Fourier transform of the corresponding latent sharp image patch.  For
whole image restoration, the input image is cut into overlapping
patches, each of which is independently processed by the network.  The
outputs are recomposed to yield an initial estimate of the latent
sharp image, which is then used together with the blurry input image
for the estimation of a single invariant blur kernel. The final result
is obtained using the state-of-the-art non-blind deconvolution method
of Zoran and Weiss \cite{zoran2011learning}.

\textbf{Multi-frame blind deconvolution.}  Splitting an exposure
budget across many photos can lead to a significant quality advantage
\cite{hasinoff2009time}. Thus, it has been shown that multiple
captured images can help alleviating the illposedness of the BD problem
\cite{rav2005two}. This has been exploited in several approaches
\cite{sparseprior,chen2008robust,cai2009blind,vsroubek2012robust,zhu2012deconvolving}
for \emph{multi-frame} BD, i.e. combining multiple, differently
blurred images into a single latent sharp image. Generative methods,
that make explicit use of an image formation model, mainly differ in
the prior and/or the optimization procedure they use. State-of-the-art
methods use sparse priors with fast Bregman splitting techniques for
optimization \cite{cai2009blind}, or within a variational inference
framework \cite{sparseprior}, cross-blur penalty functions between image
pairs \cite{vsroubek2012robust,zhu2012deconvolving}, also in
combination with robust cost functions \cite{chen2008robust}.

More recently proposed methods
\cite{zhang2014multi,zhang2015intra,kim2016dynamic,ito2014blurburst}
also model the inter-frame motion and try to exploit the interrelation
between camera motion, blur and image mis-alignment. Camera
motion, when integrated during the exposure time of a single frame
will produce intra-frame motion blur while leading to inter-frame
mis-alignment during readout time between consecutive image capture.

All of the above-mentioned methods employ generative models and try to
explicitly estimate one unknown blur kernel for each blurry input
frame along with predicting the latent sharp image. A common
shortcoming is the large computational burden with typical
computation times in the order of tens of minutes, which hinders their
wide-spread use in practice.

Recently, Delbracio and Sapiro have presented a fast method that
aggregates a burst of images into a single image that is both sharper
and less noisy than all the images in the burst \cite{fba}. The
approach is inspired by a recently proposed Lucky Imaging method
\cite{garrel2012highly} targeted for astronomical imaging. Traditional
Lucky Imaging approaches would select only a few ``lucky'' frames from
a stack of hundreds to thousands recorded short-exposure images and
combine them via non-rigid shift-and-add techniques into a single
sharp image. In contrast, the authors of \cite{garrel2012highly}
propose to take all images into account (rather than a small subset of
carefully chosen frames) taking and combining what is less blurred of each frame
to form an improved image. It does so by computing a weighted average
of the Fourier coefficients of the registered images in the burst.  In
\cite{fba,delbracio2015hand} it has been demonstrated that this
approach can be adapted successfully to remove camera shake originated
from hand tremor vibrations. Their Fourier Burst Accumulation (FBA)
approach allows for fast processing even for Megapixel images while at
the same time yielding high-quality results provided that a ``lucky'',
i.e. almost sharp frame is amongst the captured image burst. 

In our work, we not only present a learning-based variant of FBA but
also show how to alleviate the drawback of requiring a sharp frame
amongst the input sequence of images. To this end we combine the
single image BD method of \cite{chakrabarti} with FBA in a single
network architecture which facilitates end-to-end learning. To the
best of our knowledge this is the first time that a fully
discriminative approach has been presented for the challenging problem
of multi-frame BD.


\section{Method}
\label{sec:method}
Let's assume we have given a burst of observed color images
$Y_1, Y_2,\ldots, Y_N \in \mathcal{I}$ capturing the same scene
$X \in \mathcal{I}$. 
Assuming each image in the captured sequence is blurred differently,
our image formation model reads
\begin{equation}
  Y_t = k_t* X + \varepsilon_t,
  \label{eq:model_t}
\end{equation}
where $*$ denotes the convolution operator, $k_t$ the blur kernel for
observation $Y_t$ and $\varepsilon_t$ additive zero-mean Gaussian noise.


We aim at predicting a latent single sharp image $\hat{X}$
through a deep neural network architecture, i.e.
\begin{align*}
  \pi^{(\theta)}\colon \mathcal{I}_p^N \to \mathcal{I}_p,\qquad (y_1,y_2,\ldots,y_N)\mapsto \hat{x} = \pi^{(\theta)}(y_1,y_2,\ldots,y_N).
\end{align*}
The network operates on a patch-by-patch basis, here $y_t \in
\mathcal{I}_p$ and $\hat{x} \in \mathcal{I}_p$ denote a patch in $Y_t$
and $X$ respectively. The patches are chosen to be overlapping. Our
network predicts a single sharp patch $\hat{x}\in \mathcal{I}_p$
from multiple input patches $y_t\in \mathcal{I}_p$. All predicted
patches are recomposed to form the final prediction $\hat{X}$ by averaging the predicted pixel values.  During
the training phase we optimize the learning parameters $\theta$ by
directly minimizing the objective
\begin{equation}
  \Vert \pi^{(\theta)}(y_1,y_2,\ldots,y_N) - x \Vert_2^2.
  \label{eq:objective}
\end{equation}
In the following we will describe the construction of
$\pi^{(\theta)}(\cdot)$, the optimization of network parameters
$\theta$ during the training of the neural network and the restoration
of an entire sharp image.

\subsection{Network Architecture}

\begin{figure}[h!]
  \centering
  \includegraphics{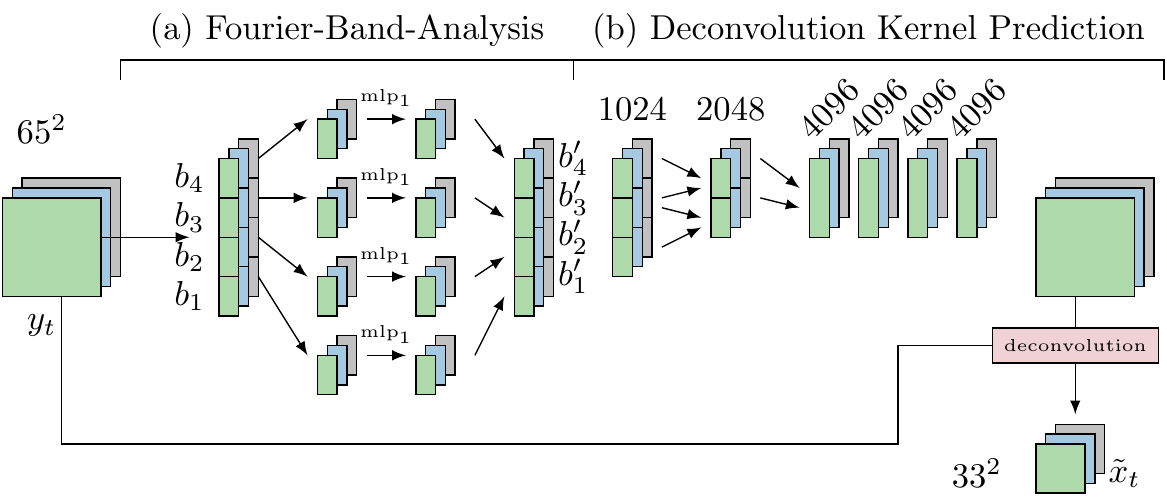}
  \caption{Frequency band analysis and deconvolution for an image
    burst with 3 patches $y_1,y_2,y_3$. Following the work of
    Chakrabarti \cite{chakrabarti} we separate the Fourier spectrum in
    4 different bands $b_1,\ldots,b_4$. In addition, we allow each band separately to interact across all images in one burst to support early information sharing. The predicted output of the deconvolution step are smaller patches $\tilde{x_1},\tilde{x_2},\tilde{x_3}$.}
  \label{fig:fourier_band_analysis}
\end{figure}

The architecture $\pi^{(\theta)}(\cdot)$ consists of several stages:
(a) frequency band analysis with Fourier coefficient prediction, (b) a
deconvolution part and (c) image
fusion. Figure~\ref{fig:fourier_band_analysis} illustrates the first
two stages of our proposed system.

\textbf{(a) Frequency band analysis.} The frequency band analysis
computes the discrete Fourier transform of the observed patch $y_t$
according to the neural network approach in \cite{chakrabarti} at
three different sizes ($17\times 17, 33\times 33, 65\times 65$) using
different sample sizes, which we will refer to bands $b_1, b_2,
b_3$. In addition, band $b_4$ represents a low-pass band containing
all coefficients with $\max \abs{z}\leq 4$ from band $b_3$. This is depicted
in Fig.~\ref{fig:fourier_band_analysis}.  To enable early information
sharing within one burst of patches, we allow the neural network to
spread the per band information extracted from one patch across all
images of the burst using $1\times 1$ convolution.

This essentially embeds a fully connected neural network for each
Fourier coefficient $(f_{ij})_{t}$ with weight sharing. Since we use
these operations in the image fusion stage again, we elaborate on this idea in more detail.

\begin{figure}[h!]
  \centering
  \includegraphics{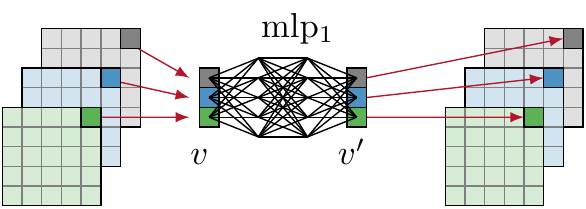}
  \caption{For arbitrary inputs (bands $b_1,b_2,b_3,b_4$ or later FBA
    weights) we interpret each coefficient across one burst as a
    single vector. A transformed version of this excerpt will be
    placed at the same location in the output patch again. To reduced the
    number of learnable parameters, we employ weight sharing
    independent of the position. }
  \label{fig:conv11}
\end{figure}

The values of one Fourier coefficient $(f_{ij})_{t}$ at frequency
position $(i,j)$ across the entire burst $t=1,2,\ldots,N$ can be
considered as a single vector $(f_{ij})_{t=1,2,\ldots,N}$ of dimension
$N$ (compare Fig.~\ref{fig:conv11}).
Each of these vectors is fed through a small network of fully
connected layers, labeled by \texttt{mlp\_1} in
Fig.~\ref{fig:fourier_band_analysis}. This allows the neural network
to adjust the extracted Fourier coefficients right before a
dimensionality reduction occurs. These modified values
$(f'_{ij})_{t=1,2,\ldots,N}$ give rise to adjusted Fourier bands $b_1', b_2',
b_3', b_4'$.

\textbf{(b) Deconvolution.} Pairwise merging of the resulting bands
$b_1', b_2', b_3', b_4'$ with modified Fourier coefficients using
fully connected layers with ReLU activation units entails a
dimensionality reduction. The produced 4096 feature vector encoding is
then fed through several fully connected layers producing a 4225
dimensional prediction of the filter coefficients of the deconvolution
kernel. Applying the deconvolution kernel predicts a sharp patch
$\hat{x}$ of size $33\times 33$ from each input sequence of
patches. This step is implemented as a multiplication of the predicted
Wiener Filter with the Fourier transform of the input patch.

\textbf{(c) Image fusion.} In the last part of our pipeline we fuse all available sharp patches
$y_1,y_2,\ldots,y_N$ by adopting the FBA approach described in
\cite{fba} as a neural network component with learnable weights.  The
vanilla FBA algorithm applies the following weighted sum to a Fourier
transform $\hat{\alpha}$ of a patch $\alpha$:
\begin{align}
  u(\hat{\alpha}) &= \mathcal{F}^{-1}\pa{\sum_{i=1}^N w_i(\zeta) \hat{\alpha}_i(\zeta)}(x) \label{eq:fba}\\
  w_i(\zeta) &= \frac{\abs{\hat{\alpha}_i(\zeta)}^p}{\sum_{j=1}^N \abs{\hat{\alpha}_j(\zeta)}^p},
\end{align}
where $w_i$ denotes the contribution of frequency $\zeta$ of a patch
$\alpha_i$.  Note, that $u(\hat{\alpha})$ is differentiable in
$\hat{\alpha}$ allowing to pass gradient information to previous layers through
back-propagation. To incorporate this algorithm as a neural
network layer into our pipeline, we replace Equation \eqref{eq:fba}
by a parametrized version
\begin{align}
  u(\hat{\alpha}) &= \mathcal{F}^{-1}\pa{\sum_{i=1}^N h_\phi(\zeta) \hat{\alpha}(\zeta)}(x).
\end{align}
Hence, instead of a hard-coded weight-averaging (using $w_i$) the
network is able to learn a data-dependent weighted-averaging scheme. Again, the function
$h_\phi(\cdot)$ represents two 1x1 convolutional layers with trainable
parameters $\phi$ following the same idea of considering the Fourier
coefficient across one burst as a single vector (compare
Fig.~\ref{fig:conv11}).


\subsection{Training} 
\begin{figure}[!ht]
  \centering
  \includegraphics[scale=1]{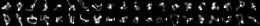}
  \caption{Some of the synthetically generated PSFs using a Gaussian process for generating training examples on-the-fly.}
  \label{fig:artificial_psf}
\end{figure}

The network is trained on an artifically
generated dataset obtained by applying synthetic blur kernels to
patches extracted from the MS COCO dataset \cite{mscoco}. This dataset
consists of real-world photographs collected from the internet. To
increase the quality of ground-truth patches guiding the training
process we reject patches with too small image gradients. This process
gives us 542217 sharp patches. For a fair evaluation we use a splitting\footnote{Provided by \cite{mscoco}} in training and validation set. Optimizing the
neural network parameters is done on the training set only.  The input
bursts of 14 blurry images are generated on-the-fly by applying
synthetic blur kernels to the ground-truth patches. These synthetic
blur kernels of sizes $17\times 17$ and $7\times 7$ pixels are
generated using a Gaussian process with a Mat\'{e}rn covariance
function following \cite{Schuler_PAMI15}, a random subset of which is
shown in Fig.~\ref{fig:artificial_psf}. In addition, we apply standard
data augmentation methods like rotating and mirroring to the
ground-truth data. Hence, this approach gives nearly an infinite
amount of training data. We also add zero-mean Gaussian noise with
variance 0.1.  The validation data is precomputed to ensure fair
evaluation during training.

Unfortunately, sophisticated stepsize heuristics like Adam
\cite{adamsolver} or Adagrad \cite{adagrad} failed to guarantee a
stable training. We suspect the large range of values in the Fourier
space to mislead those heuristics. Instead, we use stochastic
gradient descent with momentum ($\beta=0.9$), batchsize 32 and an
initial learning rate of $\eta=2$ which decreases every 5000 steps by
a factor of $0.8$. Training the neural network took 6 days using
TensorFlow \cite{tensorflow} on a NVIDIA Titan X.

The FBA approach \cite{fba} applies Gaussian smoothing to the weights
$w_i$ to account for the fact that small camera shakes are likely to
vary the Fourier spectrum in a smooth way. While this removes strong
artefacts in the restored recomposed image, it prevents the network to
convergence during training. Following this idea we tried a fixed
Gaussion blur with parameters set to the reported values of \cite{fba}
as well as learning a blur kernel (initialized by a Gaussian) during
training. In both cases we observed no convergence during training. 
Therefore, we apply this smoothing only for the final application of the
neural network.

\subsection{Deployment}
\label{subsec:deploy}
During deployment we feed input patches of size $65 \times 65$ into
our neural network with stride 5. Using overlapping patches helps to
average multiple predictions.  For recombination of overlapping
patches we apply a 2-dimensional Hanning window to each patch to
favour pixel values in the patch center and devaluate information at
the border of the patch.

\begin{figure}[h]
  \centering
  \includegraphics{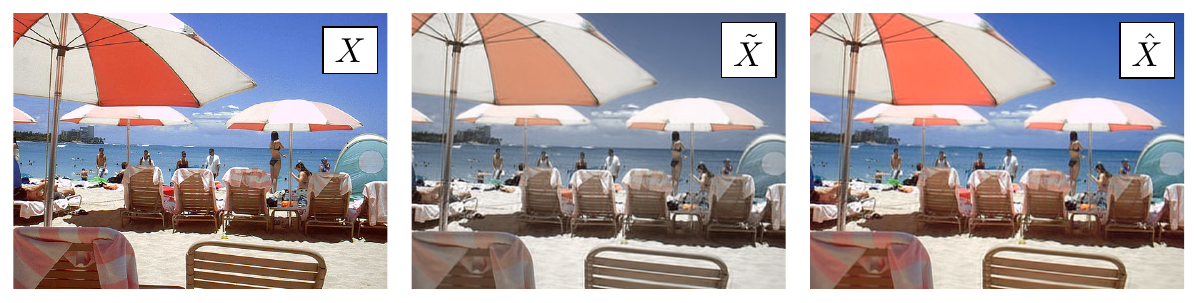}
  \caption{Deblurring a burst of degraded images from a groundtruth image (left) results in a desaturated image (middle). Therefore we correct those colors (right image) using color transfer.}
  \label{fig:color_correction}
\end{figure}

While the predicted images $\hat{X}$ generated by our neural network
contain well-defined sharp edges we observed desaturation in color
contrast. To correct the color of the predicted image we replace its
$ab$-channel in the $Lab$ color space by the $ab$-channel of the FBA
results (compare Figure~\ref{fig:color_correction}).

Regarding runtime the most expensive step is the frequency band
analysis. Given a burst of 14 images of size $1000 \times 700$ pixels
the entire reconstruction process takes roughly 5 minutes per channel
with our unoptimized implementation.

\section{Experiments}
\label{sec:results}
To evaluate and validate our approach we conduct several experiments
including a comprehensive comparison with state-of-the-art techniques
on a real-world dataset, and a performance evaluation on a synthetic
dataset to test the robustness of our approach with varying
image quality of the input sequence.

\subsection{Comparison on real-world dataset}
We compare the
restored images with other state-of-the-art multi-image blind
deconvolution algorithms. In particular, we compare with the
multichannel blind deconvolution method from \v{S}roubek et
al. \cite{vsroubek2012robust}, the sparse-prior method of
\cite{zhang2013multi} and the FBA method proposed in \cite{fba}.
We used the data provided by \cite{fba}, which contains typical
photographs captured with hand-held cameras (iPad back camera, Canon
400D). 
As they are captured under various challenging lighting conditions they
exhibit both noise and saturated pixels.
\begin{figure}[!ht]
  \centering
  \includegraphics[]{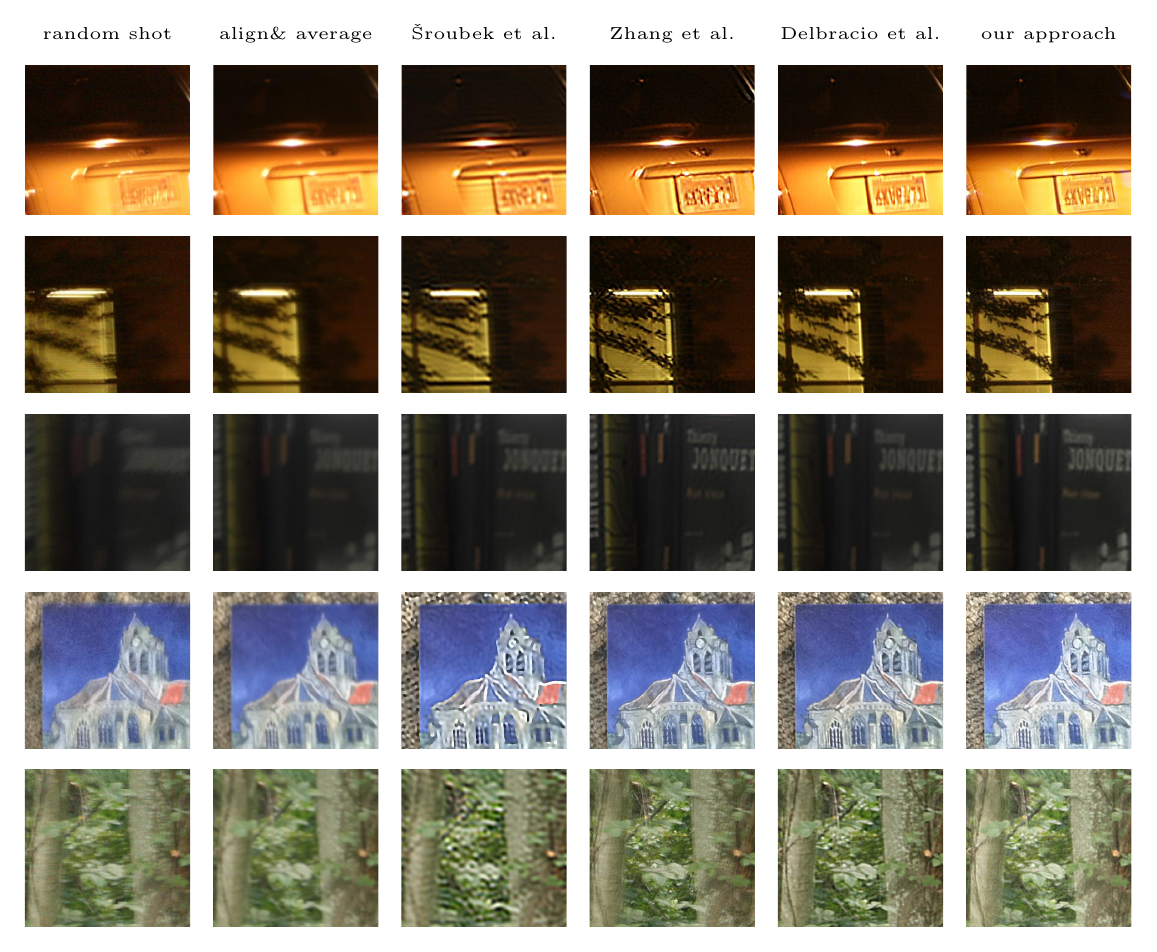}
  \caption{Comparison to state-of-the-art multi-frame blind deconvolution
    algorithms on real-world data. See the supplementary material for
    high-resolution images. Note that our approach produces the
    sharpest results except for the last scene, which could be caused
    by the color transfer described in Section~\ref{subsec:deploy}.}
  \label{fig:comparison}
\end{figure}
As shown in \cite{fba} the FBA algorithms demonstrated superior
performance compared to previous state-of-the-art multi-image blind
deconvolution algorithms \cite{vsroubek2012robust,zhang2013multi} in
both reconstruction quality and runtime. Figure~\ref{fig:comparison}
shows crops of the deblurred results on these images. The
high-resolution images are enclosed in the supplemental material.
Our trained neural network featuring the FBA-like averaging yields
comparable if not superior results compared to previous approaches
\cite{vsroubek2012robust,zhang2013multi,fba}. In direct comparison to
the FBA results, our method is better 
removing blur due to our
additional prepended deconvolution module.

\subsection{Deblurring bursts with varying number of frames and quality}
\begin{figure}[!ht]
  \centering
  \includegraphics{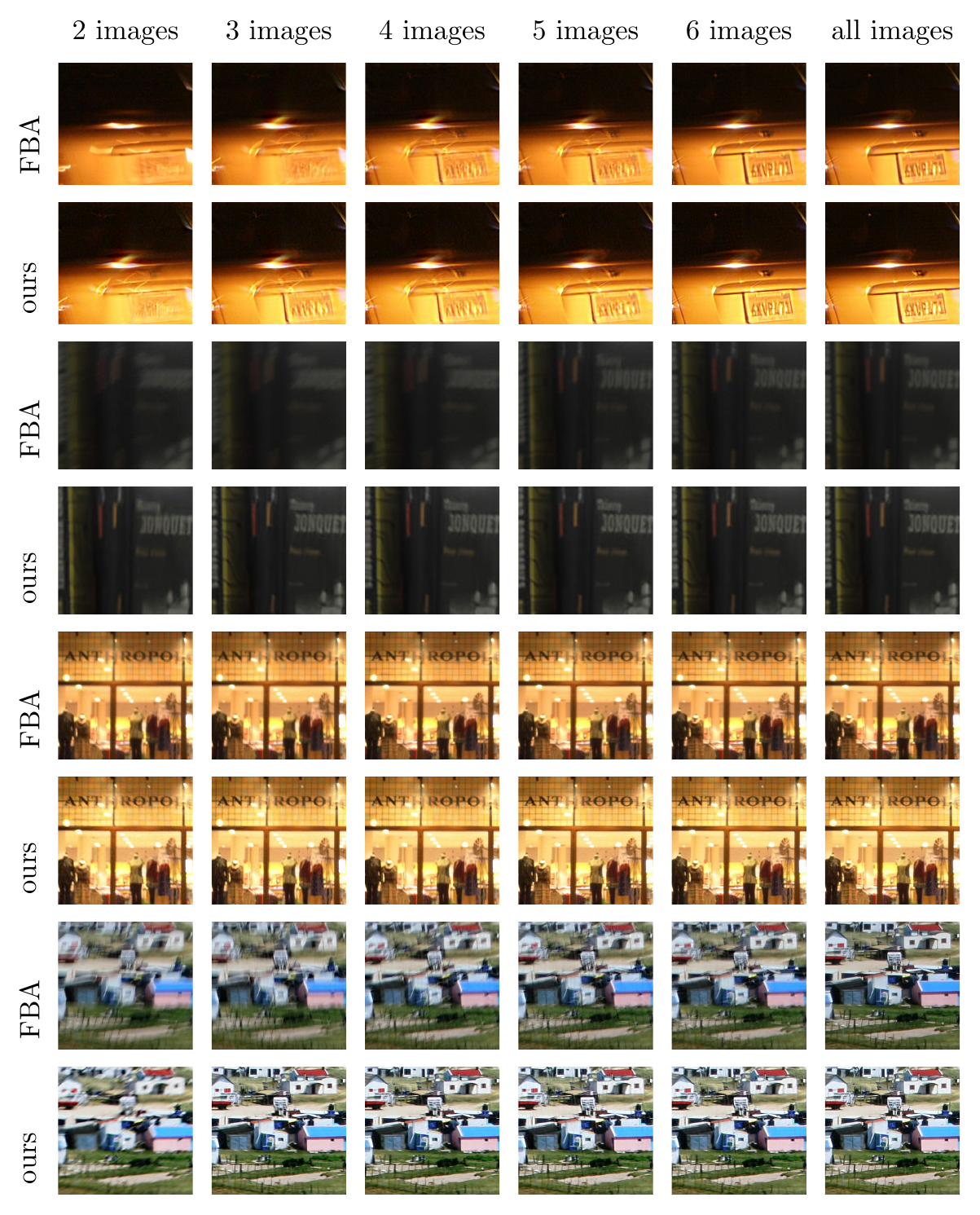}
  \caption{
FBA and our algorithm are compared on bursts with a growing number of images of increasing quality. The individual images are sorted according to their PSNR starting with the most blurry images. The input images were taken from \cite{fba}. }
  \label{fig:fba_series}
\end{figure}

Here, we analyse the performance of our approach depending on the burst
``quality''.  Sorting all images provided by \cite{fba} within one
burst according to their PSNR beginning with images of strong blur and
consequently adding sharper shots to the burst gives a series of
bursts starting with images of poor quality up to bursts with at least
one close-to-sharp shot.  Since our architecture is trained for
deblurring bursts with exactly 14 input images, we duplicated
images of bursts with fewer frames.  Figure~\ref{fig:fba_series} clearly
indicates good performance of our neural network even for a relative
small number of input images with strong blur artifact.

\subsection{Deblurring image bursts without reasonable sharp frames}
\begin{figure}[!ht]
  \centering
  \includegraphics{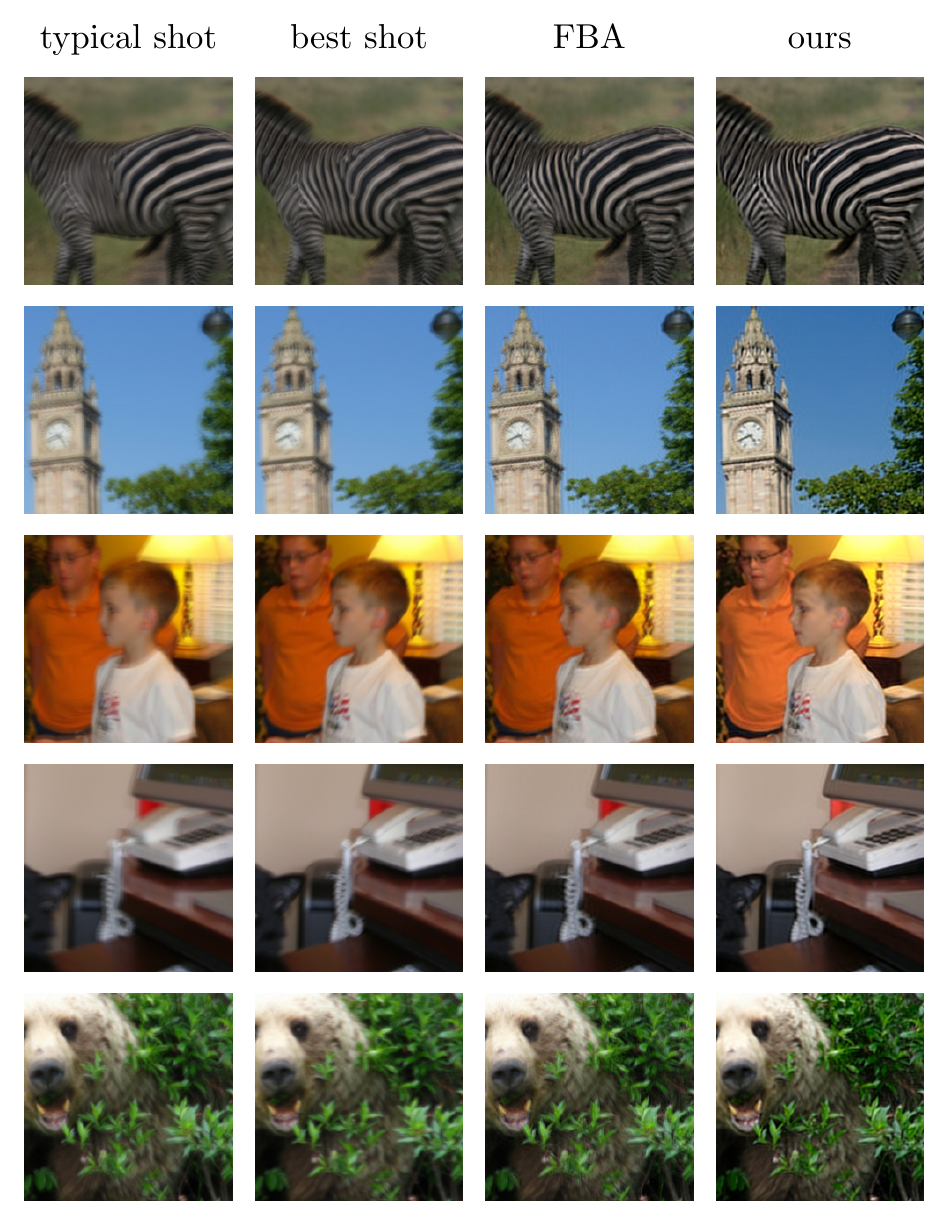}
  \caption{Comparing FBA (third column) and our trained neural network
    (fourth column) against best shot and a typical shot. These images
    are taken from the validation set. For image bursts without a
    single sharp frame lucky imaging approaches fail due to a missing
    explicit deconvolution step, while our approach gives reasonable results.}
  \label{fig:qualitative_comparison}
\end{figure}

To further challenge our neural network approach, we artificially
sampled image bursts from unseen images taken from the MS COCO
validation set and blurred them by applying synthetic blur kernels of
size $14\times 14$. The restored sharp images from the input bursts of
14 artificially blurred images under absence of a close-to-sharp
frame (best shot) are depicted in
Figure~\ref{fig:qualitative_comparison}.  As the experiments indicate
the explicit deconvolution step in our approach is absolutely
neccessary to handle these kind of snapshots and to remove blur
artifacts. In constrast, while FBA \cite{fba}
stands out in small memory footprint and fast processing times it clearly
failed to recover sharp images for cases where no reasonably
sharp frame is available amongst the input sequence.

\begin{figure}[!ht]
  \centering
  \includegraphics[width=0.9\textwidth]{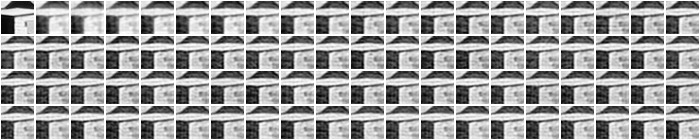}
  \caption{
  The combination of the work of Chakrabarti \cite{chakrabarti} and Delbracio et al. \cite{fba} can be considered as a baseline version of our neural network. We fine-tuned the published weights from the work of Chakrabarti \cite{chakrabarti} in an end-to-end fashion in combination with our FBA-layer. The left-most patch is the ground-truth patch. Note how the sharpness continuously increases with training. }
  \label{fig:training}
\end{figure}

\subsection{Comparing to a baseline version}
One might ask, how our trained neural network compares to an approach
that applies the methods of Chakrabarti \cite{chakrabarti} and Delbracio
and Sapiro \cite{fba} subsequently, each in a separate step. We
fine-tuned the provided weights from \cite{chakrabarti} in combination
with our FBA-layer. Figure~\ref{fig:training} shows the training
progress for an examplar patch, where the improvement in sharpness is
clearly visible.

\begin{figure}[th!]
  \centering
  \includegraphics[width=0.9\textwidth]{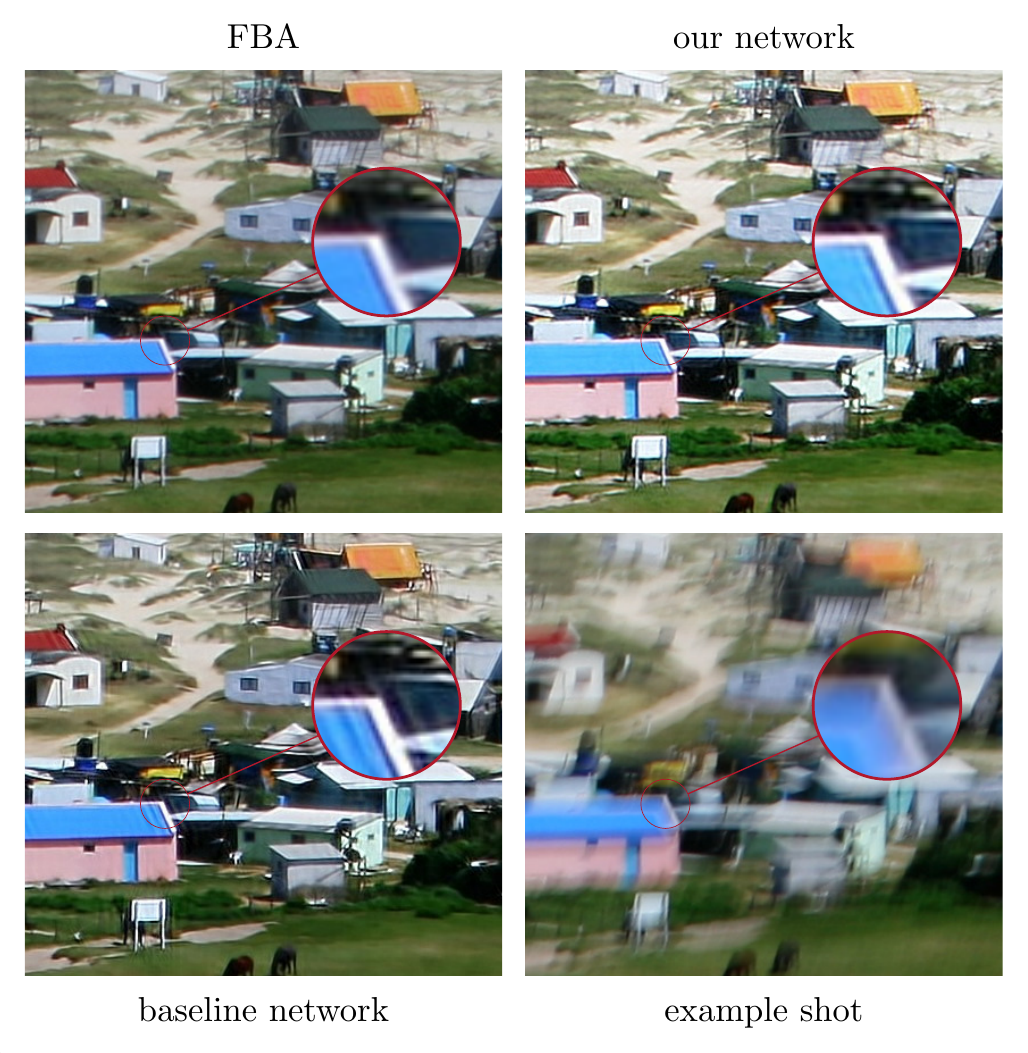}
  \caption{Comparison to a baseline approach of simply stacking \cite{chakrabarti}  and \cite{fba}. Without end-to-end training ringing-artifacts are clearly visible on the blue roof. They are significantly dampened after training.}
  \label{fig:baseline}
\end{figure}

In addition, we run the \textit{entire} pipeline of Chakrabarti
\cite{chakrabarti} including the costly non-blind deconvolution EPLL step and afterwards FBA. 
The approach is significantly slower and results in less sharp reconstructions (see Fig.~\ref{fig:baseline}).


\subsection{Spatially-varying blur}

To test whether our network is also able to deal with
spatially-varying blur we generated a burst of images degraded by
non-stationary blur. To this end, we took one of the recorded camera
trajectories of \cite{kohler2012recording} that are provided on the
project webpage\footnote{http://webdav.is.mpg.de/pixel/benchmark4camerashake/}. The
camera trajectory has been recorded with a Vicon system at 500~fps and
represents the camera motion during a slightly longer-exposed shot
(1/30s). The trajectory comprises a 6-dimensional time series with 167
time samples. We divided this time series into 8 fragments of
approximately equal lengths. 

\begin{figure}[th!]
  \centering
  \includegraphics[width=0.9\textwidth]{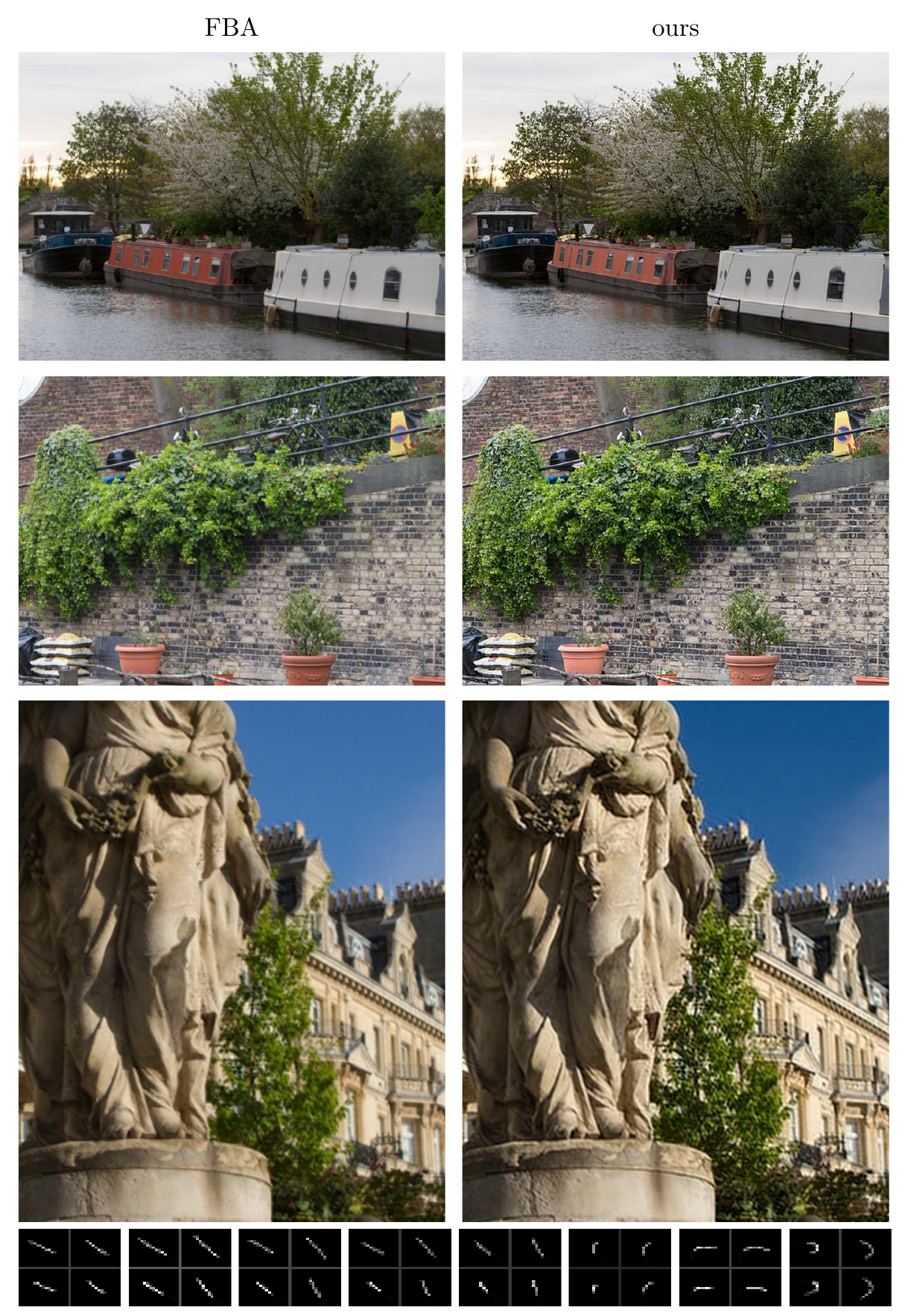}
  \caption{Comparison to FBA on image sequences with spatially-varying
    blur. Our approach is able to reconstruct consistently sharper images.}
  \label{fig:varyingblur}
\end{figure}

With a Matlab script (see Supplemental material) 8 spatially-varying
PSFs are generated as shown at the bottom of Fig. \ref{fig:varyingblur}. 

\newpage 
The spatially varying kernels of size 17 $\times$ 17 pixels are applied using the Efficient Filter Flow model of \cite{hirsch2010efficient}. The results of our network
along with results of FBA for three example images are shown in Fig.~\ref{fig:varyingblur}. Our results are consistently sharper and
demonstrate that our approach is also able to correct for
spatially-varying blur.

\section{Conclusion, Limitations and Future Work}
\label{sec:conclusion}
\vspace{-2mm}
We presented a discriminative approach for multi-frame blind
deconvolution (BD) by posing it as a nonlinear regression
problem. As a function approximator, we use a deep layered neural
network, whose optimal parameters are learned from artificially
generated data. Our proposed network architecture draws inspiration
from two recent works as (a) a neural network approach to single image
blind deconvolution of Chakrabarti \cite{chakrabarti}, and (b) the
Fourier Burst Accumulation (FBA) algorithm of Delbracio and
Sapiro \cite{fba}. The latter takes a burst of images as input and
combines them through a weighted average in the frequency domain to a
single sharp image. We reformulated FBA as a learning method and
casted it into a deep layered neural network. Instead of resorting to
heuristics and hand-tuned parameters for weight computation, we learn
optimal weights as network parameters through end-to-end
training.

By prepending parts of the network of Chakrabarti to our FBA network
we are able to extend its applicability by alleviating the necessity
of a close-to-sharp frame being amongst the image burst. Our system is
trained end-to-end on a set of artificially generated training
examples, enabling competitive performance in multi-frame BD, both
with respect to quality and runtime.  Due to its novel information
sharing in the frequency band analysis stage and its explicit
deconvolution step, our network outperforms state-of-the-art
techniques like FBA \cite{fba} especially for bursts with few severely
degraded images.

Our contribution resides at the experimental level and despite
competitive results with state-of-the-art, our proposed
approach is subject to a number of limitations. However, at the same
time it opens up several exciting directions for future research:\\[-3mm]
\begin{itemize}
\item Our proposed approach doesn't exploit the temporal structure of
  the input image sequence, which encodes valuable information about
  intra-frame blur and inter-frame image mis-alignment
  \cite{zhang2015intra,kim2016dynamic,ito2014blurburst}. Embedding our
  described network into a network architecture akin to the
  spatio-temporal auto-encoder of P\u{a}tr\u{a}ucean et
  al. \cite{patraucean2015spatio} might enable such non-trivial
  inference.

\item Our current model assumes a static scence and is not able to
  handle object motion. Inserting a Spatial Transformer Network Layer
  \cite{jaderberg2015spatial} which also facilitates optical flow
  estimation \cite{patraucean2015spatio} could be an interesting
  avenue to capture and correct for object motion occuring between
  consecutive frames.




\end{itemize}

\vspace{3mm}
\noindent {\bf Acknowledgement}. This work has been partially supported by the DFG Emmy Noether fellowship Le 1341/1-1 and an NVIDIA hardware grant. 
\newpage
\bibliographystyle{splncs}
\bibliography{references}


\end{document}